\documentclass[conference]{IEEEtran}
\IEEEoverridecommandlockouts
\usepackage{cite}
\usepackage{amsmath,amssymb,amsfonts}
\usepackage{algorithmic}
\usepackage{graphicx}
\usepackage{textcomp}
\usepackage{xcolor}
\usepackage{booktabs}
\usepackage{subcaption}
\usepackage{hyperref}
\usepackage{tikz,pgfplots}

\def\BibTeX{{\rm B\kern-.05em{\sc i\kern-.025em b}\kern-.08em
    T\kern-.1667em\lower.7ex\hbox{E}\kern-.125emX}}
\begin{document}

\title{
MCEL: Margin-Based Cross-Entropy Loss for Error-Tolerant Quantized Neural Networks
}
\author{
\IEEEauthorblockN{
Mikail Yayla and
Akash Kumar
}

\IEEEauthorblockA{Ruhr-Universität Bochum, Germany}
\IEEEauthorblockA{\{mikail.yayla, akash.kumar\}@rub.de}
}

\maketitle

\begin{abstract}
Robustness to bit errors is a key requirement for the reliable use of neural networks (NNs) on emerging approximate computing platforms and error-prone memory technologies.
A common approach to achieve bit error tolerance in NNs is injecting bit flips during training according to a predefined error model.
While effective in certain scenarios, training-time bit flip injection introduces substantial computational overhead, often degrades inference accuracy at high error rates, and scales poorly for larger NN architectures.
These limitations make error injection an increasingly impractical solution for ensuring robustness on future approximate computing platforms and error-prone memory technologies.

In this work, we adopt a fundamentally different perspective and investigate the mechanisms that enable NNs to tolerate bit errors without relying on error-aware training.
We establish a direct connection between bit error tolerance and classification margins at the output layer.
Building on this insight, we propose a novel loss function, the {Margin Cross-Entropy Loss} (MCEL), which explicitly promotes logit-level margin separation while preserving the favorable optimization properties of the standard cross-entropy loss.
Furthermore, MCEL
introduces an interpretable margin parameter that allows robustness to be tuned in a principled manner.
Extensive experimental evaluations across multiple datasets of varying complexity, diverse NN architectures, and a range of quantization schemes demonstrate that MCEL substantially improves bit error tolerance, up to 15\% in accuracy for an error rate of 1\%.
Our proposed MCEL method is simple to implement, efficient, and can be integrated as a drop-in replacement for standard CEL.
It provides a scalable and principled alternative to training-time bit flip injection, offering new insights into the origins of NN robustness and enabling more efficient deployment on approximate computing and memory systems.
\end{abstract}




\section{Introduction}


Recent studies on efficient neural network (NN) inference systems have increasingly explored the use of approximate memory and approximate computing units to reduce energy consumption, latency, and hardware complexity.
In approximate memory systems, such approximations are commonly realized by lowering the supply voltage or tightening timing parameters, thereby enabling lower power consumption or faster memory access.
When these techniques are pushed aggressively, however, they can lead to high bit error rates (BERs).
This behavior has been observed in both volatile memories, such as SRAM~\cite{reagan:2016,yang:2017} and DRAM~\cite{koppula:2019,wan:2024}, as well as emerging non-volatile memories, including STT-RAM~\cite{pham:2022}, RRAM~\cite{aziza:2023,zhang-rram:2020,hirtzlin-rram:2019}, and FeFET~\cite{yayla-fefet:2022}.
In all of these technologies, the BER increases steeply as the supply voltage is reduced and more aggressive timing constraints are imposed on the memories.

Beyond memory, approximate computing units have also been proposed as an effective means to improve the efficiency of NN inference.
A comprehensive overview of approximation techniques across different abstraction layers is provided in~\cite{mittal:2016}.
For instance, multiply-accumulate (MAC) units can be tuned to trade computational accuracy for reduced resource usage, or selected computations can be skipped to improve efficiency~\cite{armeniakos:2022}.
Other approaches exploit HW/SW co-design to select circuits with suitable approximation levels that enable NN inference with minimal impact on accuracy~\cite{sahoo:2025}.
Furthermore, analog-computing-based hardware, which has been shown to offer high energy efficiency~\cite{chi:2016,shafiee:2018}, naturally aligns with approximate computing, as signal representations and computations in the analog domain are inherently imprecise.
Without appropriate countermeasures, however, bit errors arising from approximate memory or approximate computation can degrade NN accuracy to unacceptable levels.

A common approach to improve bit error tolerance in NNs is to inject bit flips during training according to a predefined error model.
While effective for achieving robustness under small error rates, training with bit flip injection exhibits several important drawbacks.
First, recent studies have shown that injecting bit errors during training can significantly degrade inference accuracy, particularly as the bit error rate applied during training increases~\cite{hirtzlin-rram:2019,koppula:2019,buschjaeger-towards:2020}.
Second, bit flip injection introduces substantial computational overhead~\cite{mrazek:2019}.
During training, a stochastic decision must be made for every bit of error-prone data to determine whether a bit flip occurs, resulting in a large number of additional operations and increased training complexity.
While the penalty varies with error model complexity, even simple models can increase simulation time considerably.
For example, the approach in~\cite{yayla-lim:2022} requires replacing the native PyTorch MAC engine to facilitate error injection for custom XNOR-logic-in-memory circuits, thereby increasing simulation time by an order of magnitude.
Finally, as NN models, including those targeting embedded and edge systems, continue to grow in size and complexity, training-time error injection becomes increasingly difficult to scale.
This challenge is further exacerbated when bit flip injection is combined with techniques targeted for embedded NNs, such as quantization-aware training or knowledge distillation.
As a result, training-time bit-flip injection is unlikely to represent a sustainable long-term solution for achieving robustness on future approximate computing platforms and other reliability-constrained systems.

Achieving bit error tolerance in NNs without relying on bit flip injection would eliminate these limitations and constitute a significant advancement for NN deployment on approximate hardware and error-prone memory.
However, realizing this goal requires a deeper understanding of the fundamental mechanisms that enable NNs to tolerate bit errors.
To date, these mechanisms have not been systematically investigated.
While initial insights have been reported for binarized neural networks (BNNs)~\cite{buschjaeger:2021,buschjaeger-betm:2021}, corresponding analyses for generic quantized neural networks (QNNs) remain largely unexplored.
This gap is particularly relevant, as QNNs are widely adopted in practical applications and deployed across a broad range of real-world systems.
Moreover, the multi-bit nature of QNNs fundamentally alters how bit errors propagate through the network, limiting the direct transferability of insights obtained.

In this work, we close this gap and, to the best of our knowledge, present the first method to optimize QNNs for bit error tolerance without employing bit flip injection during training.
We demonstrate that bit error tolerance can be achieved as a direct consequence of enforcing margin properties in the output of the classification layer, rather than through exposure to the error model during training.

\begin{figure}[t]
    \centering
        \begin{tikzpicture}[scale=1.0]
            \draw[->, thick] (0,0) -- (0,3) node[right] {Logit};
            \draw[->, thick] (0,0) -- (6.2,0) node[below] {Class index};
            
            \draw[dashed, thick, violet] (0.3,2.5) -- (6,2.5);
            \draw[dashed, thick, violet] (0.3,1.5) -- (6,1.5);
            
            
            \foreach \x/\h in {1/2.5, 2/1.5, 3/0.9, 4/1.1, 5/1.2} {
                \draw[thick] (\x,0) -- (\x,\h);
                \node at (\x,\h) {$\times$};
            }
            
            \foreach \x/\lbl in {1/0, 2/1, 3/2, 4/3, 5/4} {
                \node[below] at (\x,0) {\lbl};
            }
            
            \draw[<->, thick, violet] (3.3,2.5) -- (3.3,1.5);
            \node[right, violet] at (3.25,2) {${m}$};
        \end{tikzpicture}
    \caption{Illustration of output-layer logits with a margin $m$. Five classes are shown as an example. Class 0 is the prediction and the extent of the margin to the second highest logit determines the error tolerance of the NN.}
    \label{fig:logit-margin-illustration}
\end{figure}
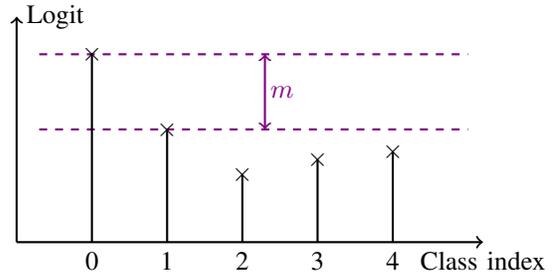

\vskip 0.1in
\noindent \textbf{Our contributions:} 
\begin{itemize}
    \item We establish a direct connection between bit error tolerance in NNs and output-layer margins.
    In particular, we show that the difference between the largest and second-largest output logits governs the NN's ability to tolerate parameter perturbations (see Figure~\ref{fig:logit-margin-illustration}).
    
    \item Based on this insight, we derive a novel loss function, the {Margin Cross-Entropy-Loss (MCEL)}, which explicitly incorporates margins into cross-entropy-based optimization. 
    The formulation is interpretable and exposes a single design parameter that allows practitioners to directly control the targeted level of error tolerance.
    
    \item We demonstrate that MCEL substantially improves bit error tolerance 
    across a wide range of scenarios.
    Our evaluation spans datasets of various complexity (FashionMNIST, SVHN, CIFAR10, and Imagenette), diverse network architectures (VGG3, VGG7, MobileNetV2, and ResNet18), and multiple quantization schemes (binary, 2-, 4-, and 8-bit).
    Our method is simple to implement, efficient, and can be integrated as a drop-in replacement for standard CEL.
    The source code of our framework is available at \url{https://github.com/myay/BNN-QNN-ErrorEvaluation}.
\end{itemize}

\section{Related Work}

Prior work has investigated the robustness of NNs for deployment on approximate memory or approximate computing units.
Across these studies, robustness is typically achieved in one of two ways: Either the NN exhibits an inherent tolerance to hardware-induced errors, or error-aware training techniques are employed such that the NN maintains high inference accuracy even in the presence of errors.

\paragraph{Error-tolerant NNs on approximate memory}
For approximate DRAM-based systems, the study in~\cite{koppula:2019} provides a comprehensive overview of NN inference using different DRAM technologies and propose a framework for evaluating NN accuracy under approximate DRAM operation across various system configurations.
Their study demonstrates that DRAM parameters, such as supply voltage and timing margins, can be tuned to significantly improve energy efficiency and performance, while the resulting accuracy degradation remains negligible when retraining is applied.
Beyond DRAM, emerging non-volatile memories (NVMs), such as STT-RAM~\cite{hirtzlin-mram:2019}, RRAM~\cite{hirtzlin-rram:2019}, and FeFET~\cite{yayla-fefet:2022} have also been investigated in the context of approximate NN inference.
In these technologies, approximation techniques such as low-voltage reads or writes are employed to reduce energy consumption or extend device endurance.
While effective, these approaches typically rely on retraining or error-aware optimization to compensate for the increased bit error rates introduced by approximate memory operation.

\paragraph{Error-tolerant NNs for approximate computing units}
Error tolerance has also been studied extensively for approximate computing units~\cite{mittal:2016}. 
Approximate arithmetic units, such as multiply–accumulate (MAC) units with reduced precision or relaxed timing, trade computational accuracy for reduced resource usage and improved efficiency~\cite{armeniakos:2022}.
Several studies demonstrate that NNs can tolerate such approximations when trained accordingly.
In many cases, retraining or architecture-specific adaptations are required to preserve inference accuracy under approximate computation~\cite{sahoo:2025}.

\paragraph{Fault- and error-tolerant NNs}
A broader perspective on fault- and error-tolerant NNs is provided by the survey in~\cite{huitzil:2017}, which reviews both early and more recent approaches to improving NN robustness.
Representative examples include the work in~\cite{edwards:1997}, which introduces penalty terms to encourage an optimal distribution of computation across neurons to enhance fault tolerance, and the study in~\cite{cavalieri:1999}, which proposes distributing weight magnitudes evenly across neurons to reduce sensitivity to errors.
These approaches primarily focus on architectural or regularization-based robustness rather than explicitly analyzing the role of output-layer decision margins.

\paragraph{Margin-based optimization}
In parallel, margin-based concepts have been explored extensively in the context of improving class separability and discriminative power.
In particular, studies in face recognition~\cite{liu:2023,deng:2022,wang-fc:2018,wang:2018} propose incorporating margins into softmax-based loss functions to maximize inter-class variance while minimizing intra-class variance.
These methods typically enforce margins in an angular or embedding space by penalizing the target logit relative to others, thereby improving recognition performance.
While these approaches inspire our work, their primary objective is improved discrimination rather than robustness to hardware-induced perturbations.
In contrast, our work focuses on the {explicit margin between output-layer logits themselves}.
We argue that this logit-level separation directly governs the classifier’s robustness to perturbations caused by bit errors and quantization noise.
Rather than operating on angles or feature embeddings, we directly control the margin that determines the stability of the final classification decision under hardware-induced errors.

\paragraph{Relation to prior work on low-precision networks}
Margin-based robustness has previously been studied for binarized neural networks (BNNs)~\cite{buschjaeger:2021,buschjaeger-betm:2021}.
However, corresponding analyses for quantized neural networks (QNNs), where the parameters are represented as n-bit fixed point values, remain largely unexplored.
This gap is particularly significant, as QNNs are widely adopted in practical systems and deployed across a broad range of real-world applications.
Moreover, the multi-bit nature of QNNs fundamentally changes how bit errors propagate through the NNs, limiting the direct transferability of insights obtained for BNNs.
Our work addresses this gap by investigating margin-based robustness in QNNs without relying on training-time error injection.

\section{Definitions and Notations}
\label{sec:defnot}

In this work, we consider NNs that perform multiclass classification, where the predicted class is given by the index of the maximum output score.
Let $\mathcal{X} \subseteq \mathbb{R}^d$ denote the input space and let
$\mathcal{Y} = \{1, \dots, K\}$ denote the set of $K$ class labels.
A NN with parameters $\theta$ defines a function
\[
F_\theta : \mathcal{X} \to \mathbb{R}^K .
\]
For an input $x \in \mathcal{X}$, the NN output
$F_\theta(x) = (f_\theta(x)_1, \dots, f_\theta(x)_K)$ consists of $K$
class scores from an NN's output layer, where $f_\theta(x)_k$ denotes the score associated with class $k \in \mathcal{Y}$. 
For all scores we assume it holds that $f_\theta(x)_k \geq 0$.
This is without loss of generality as all scores can be shifted by a constant.
The predicted class label, i.e. the index with the highest logit, is given by
\[
\hat{y}(x) = \arg\max_{k \in \mathcal{Y}} f_\theta(x)_k .
\]

To reason about the robustness of NNs, we define the following terms. 
A bit flip is an inversion of a single bit in the binary representation of a weight or input value, i.e., a transition from 0 to 1 or from 1 to 0.
An error is defined as a transient fault manifesting as one or more bit flips in the digital representation of parameters or other data of the NN.
An error-induced misclassification occurs when the prediction produced by the NN after the occurrence of an error differs from the prediction obtained in the error-free case. 
The error tolerance of an NN is defined as its ability to preserve its prediction under perturbations to its parameters or other data. 
A more formal description is given in the following section.

\section{Relating Error Tolerance to Output-Layer Margins}
\label{sec:margins}

We propose that the ability of a NN to maintain correct predictions under parameter perturbations can be characterized by the classification margin for each input.  
We define the classification margin for an input $x \in \mathcal{X}$ as:
\begin{equation}
m(x, \theta) = f_\theta(x)_{\hat{y}(x)}
       - \max_{k \in \mathcal{Y} \setminus \{\hat{y}(x)\}} f_\theta(x)_k .
\label{eq:top2}
\end{equation}
This means that the difference $m(x, \theta)$ between the largest and second-largest output values
defines the classification margin, which quantifies the robustness of
the classification decision.
The intuition for the margin and its connection to robustness is illustrated in Figure~\ref{fig:logit-margin-illustration}.
In the case of $m(x, \theta) > 0$, the NN is defined to be error tolerant with margin $m(x, \theta)$, for the input $x$.
If $m(x, \theta) = 0$, there exists more than one class attaining the maximum
score. 
In this case, we assume that the predicted label $\hat{y}(x)$ is chosen
arbitrarily from this set of classes. 

Any perturbation of the parameters can lead to a perturbation of the margin.
Let $\delta \theta$ denote perturbations of the NN
parameters.  
With a perturbation $\delta \theta$, the resulting perturbed margin is
\[
m(x, \theta + \delta \theta).
\] 
If a perturbation $\delta \theta$ causes the classification
margin to become zero or negative, i.e.,
\[
m(x, \theta + \delta \theta) \le 0,
\]
then the perturbed parameters cause a misclassification,
i.e., the predicted label differs from the original prediction.
Note that other sources of perturbation can also affect the classification margin.
These include erroneous computations, which may be intentional (e.g. in approximate computing) or unintentional (e.g. due to errors in computations or input perturbations due to sensor noise).
Therefore, the perturbations considered here are representative and do not limit the generality of the margin-based analysis.
   
Ideally, to maximize the tolerance of NNs to perturbations,
the classification margin $m(x, \theta)$ should be as large as possible.
Consequently, during training, the NN parameters $\theta$ should
be optimized so that the margin $m(x, \theta)$ is maximized for each sample in the data set.
    
While this ensures that the NN
separates the predicted class from its nearest competitor, it does not
account for the distances to other classes, which may also be relatively
close in score. 
Consequently, the network may still be sensitive to
perturbations in the parameters that reduce the separation
to one of the other classes, leading to misclassifications.
   
To enhance robustness, we therefore consider the margin between the
predicted class and all other classes for every sample in the
dataset. 
Formally, for each sample $x_j \in \mathcal{X}$  of the dataset with $N$ samples and each
non-predicted class $k \in \mathcal{Y} \setminus \{\hat{y}(x_j)\}$, we
define the class-specific and sample-specific margin
\[
m_k(x_j, \theta) = f_\theta(x_j)_{\hat{y}(x_j)} - f_\theta(x_j)_k.
\]
This leads us to the central \textbf{problem definition of this work}: Given a set of labelled input data, the objective is to train a NN for high accuracy and high bit error tolerance by finding the NN parameters $\theta^*$ such that
\begin{equation}
\begin{aligned}
\theta^* 
&= \arg\max_{\theta} m_k(x_j, \theta), \\
&\quad \forall j \in \{1,\dots,N\}, \;
\forall k \in \mathcal{Y} \setminus \{\hat{y}(x_j)\}.
\end{aligned}
\label{eq:problemdef}
\end{equation}
Maximizing all such margins simultaneously ensures that the predicted
class is well-separated from every other class, thereby improving
tolerance to perturbations and achieving high accuracy. 
This per-class, per-sample maximization generalizes
the standard top-two margin objective.

\section{Loss Functions and Error Tolerance Optimization}
\label{sec:lossetopt}

In the following we review two popular loss functions, the hinge loss and the cross-entropy loss, and the role of the classification margin in the optimization.

\subsection{Hinge Loss}

For error tolerance maximization, the score of the predicted class $f_\theta(x_i)_{\hat{y}(x_i)}$ needs to be as large as possible, while the other $f_\theta(x_i)_{k}$ need to be as small as possible.
In the literature, the hinge loss for maximum margin classification, typically used for support vector machines (SVMs), represents a loss function that aims to achieve this objective for two classes.

The hinge loss (see \cite{rosasco:2004}), for maximum margin classification with two classes is defined as
\begin{equation}
\ell(y,f_\theta) = \max(0, 1-y \cdot f_\theta),
\label{hingeloss}
\end{equation}
with the ground truth prediction $y=\pm1$ and the prediction of the model $f_\theta \in \mathbb{R}$.
This loss becomes small if the predictions have the same sign as the predicted class and are close to 1 in magnitude. 
For predicted values larger than 1, the loss becomes 0. 
The 1 in the loss forces the classifier to maximize the margin between two class predictions. 
The hinge loss is a convex function.
Therefore, to solve optimizations problems with it, many of the common optimizers or algorithms in NNs can be used.
In NNs, stochastic gradient descent (SGD) strategy~\cite{zhang:2004} is typically used, and combined with backpropagation it is currently the standard algorithm to train NNs.

Note that the work in~\cite{buschjaeger:2021} applies a modified hinge loss (MHL) to binarized neural networks (BNNs) and demonstrates a significant improvement in bit error tolerance.
However, when MHL is transferred to quantized neural networks (QNNs), we observe a substantial degradation in classification accuracy compared to standard cross-entropy loss (CEL), and in many cases the training process fails to converge altogether.
We attribute this behavior to fundamental structural differences between BNNs and QNNs, most notably the use of binary activations and the restriction to two discrete weight levels in BNNs, as opposed to ReLU activations and multi-level weight representations in QNNs.
These differences alter both the optimization landscape and the propagation of bit errors through the network.
Consequently, rather than adapting hinge-loss-based formulations, we take cross-entropy loss (CEL), which consistently yields high accuracy for QNNs, as our starting point and extend it to promote bit error tolerance.

\subsection{Cross-Entropy Loss}

In the existing literature, a widely adopted approach to improve error tolerance in NNs is to combine the standard cross-entropy loss (CEL) with explicit bit flip injection during training.
In these works, CEL is typically used as a default choice due to its proven effectiveness for supervised classification, rather than as the result of a systematic analysis of its suitability for error-tolerant learning.
As a consequence, the role of the loss function itself in achieving robustness to bit errors has not received sufficient attention.
Instead, robustness is predominantly attributed to the presence of errors during training, while the changes that error-aware training induces in NNs remain unknown.
Motivated by this observation, we focus in this subsection on developing a principled understanding of how loss function design affects error tolerance in NNs.



To understand the CEL, we first need to cover cross-entropy. 
Intuitively, cross-entropy measures how many bits would be needed, on average, to encode events drawn from the true distribution $p$ if we used a coding scheme optimized for a different distribution $q$. 
Minimizing the cross-entropy can therefore be interpreted as minimizing the additional encoding inefficiency between $p$ and $q$.
For discrete distributions $p$ and $q$, the cross-entropy is defined as
\[
H(p,q) = -\sum_{x \in \mathcal{X}} p(x) \, \log q(x).
\]
Minimizing the cross-entropy between $p$ and $q$ corresponds to making the predicted distribution $q$ as close as possible to the true distribution $p$.
More background on the cross entropy is in~\cite{goodfellow:2016}.



In the context of NNs performing classification, the true distribution $p$ is typically represented as a one-hot vector (in the same format that also $F_\theta$ returns), where the entry corresponding to the correct class is equal to one and all other entries are zero. 
Substituting this into the cross-entropy definition simplifies the loss. 
Specifically, the summation over all classes reduces to a single term, corresponding to the ground-truth class $i$, yielding
\[
H(p,q) = -\sum_{k=1}^{K} p(k)\,\log q(\hat{y_k}) = -\log q(\hat{y_i}).
\]
where
\[
p(k) =
\begin{cases}
1, & \text{if } k = i, \\
0, & \text{otherwise},
\end{cases}
\]
and $q(\hat{y_i})$ is the probability the model assigns to class $i$.



To be able to incorporate the concept of margin into the cross-entropy formulation, the softmax function, which is applied before cross-entropy calculation, needs to be considered. 
For simplicity of presentation, we from now on refer to the NN logit of class $k$ as $\hat{y_k}$, which correspond to the values $f_{\theta}(x)_k$ as described in Sec.~\ref{sec:defnot}, and $\mathbf{\hat{y}}$ is the vector of all $\hat{{y_k}}$.
The softmax function converts the logits into a probability distribution over $K$ classes with:
\[
softmax(\mathbf{\hat{y}})_i = \frac{\exp(\hat{y_i})}{\sum_{j=1}^{K} \exp(\hat{y_j})}, \quad i = 1, \dots, K.
\]
Each  softmax output, which is in $(0,1)$, represents the predicted probability of class $i$.
With the ground truth class $i$ and softmax as input, the standard CEL is defined as
\begin{equation}
    {\ell_{CEL}(\hat{y}, i) = -\log \left( \frac{\exp(\hat{y_i})}{\sum_j \exp{(\hat{y_j})}} \right),}
    \label{eq:cel}
\end{equation}
which measures the performance of the NN prediction.
This formulation is used in the NN training procedure to compute error backpropagation and based on the results of that, update  the model such that the error between the ground truth and the NN prediction is reduced.

The CEL in Equation~\ref{eq:cel} can also be written as 
\[\log \sum_j \exp{(\hat{y_j})-\hat{y_i}}.\] 
With large differences among $\hat{y_j}$, the term $\sum_j \exp{(\hat{y_j})}$ becomes approximately the highest value $\hat{y_{j}}*$ due to the exponential, so the term ${\hat{y_{j}}*}-\hat{y_i}$ influences the optimization most (smaller logits still contribute, but less).
In case of good predictions, this term will be close to zero (because both ${\hat{y_{j}}*}$ and $\hat{y_i}$ will be close to 1).
In case of bad predictions, i.e. large $\hat{y_{j}}*$ (or large $\log \sum_j \exp{(\hat{y_j})}$ in general) and small $\hat{y_i}$, the $\hat{y_{j}}*$ will be decreased and $\hat{y_i}$ increased, to minimize the loss. 
Thus, the {CEL also leads to an increase of the margins}, as the output of the neuron with the correct class is attempted to be maximized, and the output of the neurons of the incorrect classes are pushed to become small.
Note however, that the softmax for normalization, using the exponential function, may distort the margins, making larger logits exponentially larger than smaller ones.
Unlike a hard margin maximization (e.g., in SVMs), CEL uses softer margins, since all logits influence the loss.




Overall, with this intuitive understanding, we conclude that the CEL does not directly maximize the margins as in SVMs with the hinge loss.
The CEL primarily ensures that the correct logit is larger than the incorrect ones.
All other logits still influence the loss, but their influence is decreased exponentially. 


\section{Our Novel Method: MCEL for Optimizing Error Tolerance}
\label{sec:tcm-cel}

When modifying the cross-entropy loss (CEL), it is desirable to preserve its favorable optimization properties and proven effectiveness.
Since the loss function directly determines how prediction errors are measured and how NN parameters are updated, we deliberately focus on a minimal and principled modification of CEL.
In particular, we aim to explicitly encourage a larger separation between the logit of the correct class and those of competing classes.
Related work has explored margin-based extensions of cross-entropy to improve class separability, primarily in the context of feature learning and face recognition. 
These approaches typically operate in angular or embedding spaces rather than directly on output logits~\cite{liu:2023,deng:2022,wang-fc:2018,wang:2018}.

Let $\hat{\mathbf{y}} \in \mathbb{R}^K$ denote the vector of logits produced by the NN for an input $x$, and let $i$ be the index of the ground-truth class.
A margin-boosted variant of the cross-entropy loss can be defined as
\begin{equation}
\ell_{\mathrm{CELm}}(\hat{\mathbf{y}}, i)
=
- \log \frac{\exp(\hat{y}_i - m)}
{\exp(\hat{y}_i - m) + \sum_{j \neq i} \exp(\hat{y}_j)},
\label{eq:celm}
\end{equation}
where $m > 0$ is a fixed margin applied to the ground-truth logit.
Here, we subtract $m$ from the logit of the correct class.
The idea is to thereby force the NN to produce a logit larger in magnitude than without applying $m$.
Although the predicted class may be correct and there may be a separation between the correct class logit and the other logits, we still demand a larger separation.

A fundamental issue with directly introducing such a shift $m$ into the CEL is its inherent shift invariance.
For any constant $m \in \mathbb{R}$, the softmax function satisfies
\begin{equation}
\operatorname{softmax}(\mathbf{\hat{y}}) = \operatorname{softmax}(\mathbf{\hat{y}} - m),
\label{eq:softmax}
\end{equation}
where $\mathbf{\hat{y}}$ denotes a vector of logits.
As a consequence, the NN can cheat by decreasing all logits simultaneously, instead of increasing the logit of the correct class relative to others.
This yields identical softmax probabilities to the case without shifts and a low loss value.
More generally, absolute logit magnitudes have no intrinsic meaning in softmax-based classifiers.
Only relative differences between logits affect the predicted class probabilities.

One possible countermeasure is to explicitly bound the range of admissible logit values.
By clipping logits to a fixed interval $[-L, L]$, a margin can then be enforced by subtracting a constant from the logit of the correct class.
While this approach prevents numerical overflow and limits global logit shifts, hard clipping introduces non-smooth boundaries with zero gradients outside the clipping interval.
Moreover, it relies on a fixed absolute bound and therefore lacks scale awareness.

To address the issues of unconstrained logit shifts, the absence of a meaningful reference scale for margins, and the optimization difficulties introduced by hard clipping, we propose a smooth logit clamping mechanism based on the hyperbolic tangent function.
For each logit $\hat{y}_k$, we define the clamped logit as
\begin{equation}
\tilde{y}_k
=
L \cdot \tanh\!\left(\frac{\hat{y}_k}{L}\right),
\quad k = 1,\dots,K,
\label{eq:tanhclip}
\end{equation}
where $L > 0$ is a fixed saturation bound.

This transformation behaves approximately linearly for small $|\hat{y}_k|$, i.e. it keeps the notion of margins with $L \cdot \tanh(\frac{z}{L}) \approx z$, such that $\tilde{y}_k \approx \hat{y}_k$, thereby preserving relative logit differences and classification margins for typical logit magnitudes.
By scaling the argument of the $\tanh$ function by $1/L$, the transformation closely approximates the identity function in this regime, provided that $L$ is chosen sufficiently large relative to the expected logit magnitudes.
For large $|\hat{y}_k|$ compared to $L$, the output smoothly saturates to $\pm L$, preventing uncontrolled logit growth while maintaining continuity and differentiability everywhere.
Multiplying the output of the $\tanh$ by $L$ restores the original logit scale, allowing margins to be defined directly in the logit domain.
As a result, the transformation effectively acts as a bounded, scale-preserving squashing function that leaves moderate logits unchanged and only intervenes when logits become excessively large.

\begin{figure}[!tb]
    \begin{center}
        \begin{tikzpicture}
            \begin{axis}[
                axis lines=middle,
                xmin=-4, xmax=4,
                ymin=-1.3, ymax=1.3,
                xtick=\empty,
                ytick=\empty,
                width=10cm,
                height=5cm,
                xlabel={$z$},
                ylabel={score},
                clip=false,
                xlabel style={at={(axis description cs:0.975,0.5)},anchor=north},
            ]
            
            \addplot[
                domain=-4:4,
                samples=200,
                thick,
            ]
            {tanh(x)};
            
            \addplot[
                dotted,
                thick,
                domain=-4:4
            ]
            {1};
            
            \addplot[
                dotted,
                thick,
                domain=-4:4
            ]
            {-1};
            
            \addplot[
                thick,
                violet, 
                dashed
            ]
            coordinates {(-4,0.4) (4,0.4)};
            
            \draw[<->, thick, violet]
            (axis cs:3,0.4) -- (axis cs:3,1)
            node[midway, right] {$\frac{m}{2L}$};
            
            \node at (axis cs:3.6,1.2) {$+L$};
            \node at (axis cs:3.6,-1.2) {$-L$};
            \end{axis}
        \end{tikzpicture}
    \end{center}
    \caption{Intuition behind the tanh-clipped Margin Cross-Entropy (MCEL) Loss.
    Raw logits are first passed through a scaled tanh function, bounding class scores to a finite interval $[-L, L]$ and preventing unbounded growth of logit magnitudes.
    Dotted lines: Saturation limits imposed by tanh.
    Solid curve: tanh function.
    Horizontal violet line: Competing (non-target) class score.
    The enforced margin $m$ requires the target class score to exceed competing class scores by a fixed fraction of the available dynamic range, i.e. $\frac{m}{2L}$.}
    \label{fig:tcmcel-intuition}
\end{figure}
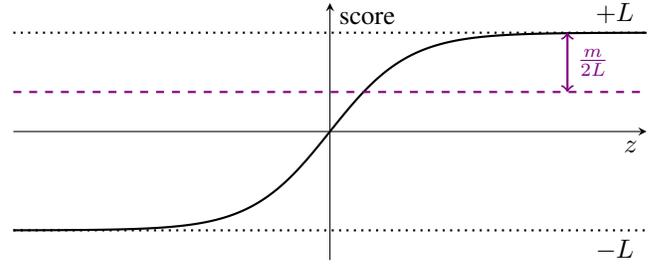

With tanh-based clamping, all logits are confined to the interval $[-L, L]$, yielding a fixed dynamic range of width $2L$.
With the confined logit interval, the margin $m$ from Equation~\ref{eq:celm} can now be reintroduced in a reasonably way. 
The margin $m$ can then be applied to enforce a minimum logit separation corresponding to a fraction
\begin{equation}
\mathrm{RLS} = \frac{m}{2L},
\label{eq:rls}
\end{equation}
which we refer to as the {Relative Logit Separation}.
For example, choosing $L = 100$ and $m = 32$ corresponds to a required separation of approximately $16\%$ of the available logit range.
This interpretation provides a clear geometric meaning for the margin and enables intuitive control over its strength.
The intuition behind this concept is shown in Figure~\ref{fig:tcmcel-intuition}.

To explicitly enforce a classification margin, we modify the clamped logit corresponding to the ground-truth class.
Let $i \in \{1,\dots,K\}$ denote the correct class index.
The margin-modified logits are defined as
\begin{equation}
\tilde{y}_i^{(m)} = \tilde{y}_i - m,
\qquad
\tilde{y}_j^{(m)} = \tilde{y}_j \quad \forall j \neq i,
\label{eq:margin-logits}
\end{equation}
where $m > 0$ is the margin parameter.
The resulting tanh-clipped margin cross-entropy loss (MCEL) is then given by

\begin{equation}
\ell_{\mathrm{MCEL}}(\hat{\mathbf{y}}, i)
=
- \log
\frac{
\exp\!\left(\tilde{y}_i - m\right)
}{
\exp\!\left(\tilde{y}_i - m\right)
+
\sum_{j \neq i} \exp\!\left(\tilde{y}_j\right)
}.
\label{eq:mcel}
\end{equation}

This formulation simultaneously ensures bounded logits, preventing degenerate solutions based on global logit shifts, explicitly enforces a margin between the correct and competing classes, and preserves meaningful logit differences for moderate values due to the approximately linear behavior of the clamping function around zero.
In contrast to hard clipping, the proposed tanh-based clamping introduces no discontinuities and maintains smooth gradients throughout training.
Moreover, by constraining logits to a predefined range, the margin parameter $m$ becomes directly interpretable through the ratio $\mathrm{RLS} = m / (2L)$, making the loss particularly well suited for scenarios in which bit error tolerance must be both tunable and interpretable.

\section{The Effect of Perturbations in QNNs/BNNs}


Up to this point, we considered NN parameters under perturbations to be in $\mathbb{R}$.
This is convenient for theoretical analyses, however, for efficient computations, these values need to be quantized.
NNs with quantized parameters, i.e. Quantized NNs (QNNs), are commonly used in everyday-life applications across various fields, where resource efficiency is an important factor. 

Before we explore the effectiveness of MCEL under bit errors, we first discuss the impact that bit flips have on the quantized NN parameters. We first cover QNNs, and then also consider BNNs, as an example of an NN architecture with a bounded (binary) activation function. 

Here we assume quantization schemes $\mathbb{R} \to Q$, where $Q$ is a set with a finite number of levels, which we denote here as ordered $Q = \{ q_1, \dots, q_L\}$, where $q_v < q_{v+1}$ for $1 \leq v \leq L$. 
If an arbitrary quantization scheme is applied to $w$, then we describe the quantized value as $Q(w) = q_v$.

\subsection{Effects of Perturbations in n-bit QNNs}

In $n$-bit QNNs, weights and activations are represented
using a finite set of discrete levels. 
Let a real-valued quantity $v$ be uniformly quantized using $n$ bits, resulting in
\[
q_l = Q(v) \in \mathcal{Q},
|\mathcal{Q}| = 2^n,
\]
where the set of possible quantization values has $2^n$ elements. 


Uniform quantization of a real-valued scalar $v \in [v_{\min}, v_{\max}]$ to an
$n$-bit integer representation can be defined as
\begin{equation}
q = \operatorname{round}\!\left(
    (v - v_{\min})
    \frac{2^{n}-1}{v_{\max}-v_{\min}}
\right).
\label{eq:quantization}
\end{equation}
Here, $v_{\min}$ and $v_{\max}$ denote the minimum and maximum values of the
quantization range. The scaling factor
$\frac{2^{n}-1}{v_{\max}-v_{\min}}$ maps the continuous value range
$[v_{\min}, v_{\max}]$ to the discrete integer interval
$\{0, \ldots, 2^{n}-1\}$. Subtracting $v_{\min}$ shifts the input range such that
$v_{\min}$ is mapped to zero, and the rounding operation yields the nearest
representable integer value.

Recall that the set of quantized values is $\mathcal{Q} = \{ q_1, \dots, q_L\}$. 
A bit flip in the encoding leads $q$ to become a value in $\mathcal{Q}$ with a different index, here $q_l$ becomes $q_{l*}$, where $l \neq l*$. 
The induced absolute error is therefore given by
\[
|q_l - q_{l \pm k}| = k \cdot \Delta,
\]
where $\Delta$ is the step size in the encoding, i.e. any two subsequent elements $q_l$ and $q_{l+1}$ (or $q_{l-1}$) have the difference $\Delta$.
The value $k$ is the number of quantization level that are crossed due to the bit flip. 
Therefore, the error scales linearly with the quantization step size. 

Note that the index for the value in $\mathcal{Q}$ is $1 \leq l \leq L$. 
Due to bit flips, values can only assume levels inside the quantization set, which prevents errors from uncontrollably propagating through the NN.
Therefore, the impact of a bit flip is bounded within the quantization set.

The value of the induced value change $k \cdot \Delta$ depends on the position of the flipped bit in the binary encoding.
Errors in more significant bits result in larger perturbations, while errors in
less significant bits induce smaller deviations.
Consider the quantized value to be an unsigned integer of the form $B_{b-1} \dots B_{1} B_{0}$, where $B_i$ can be 1 or 0.
The bit flip with the largest impact is at the position of the most significant bit, here it is $B_{b-1}$. 
This means the impact of one bit flip can be bounded by 
\[
k \cdot \Delta \leq 2^{b-1}.
\]
The position with the second most impact is at $b-2$, which follows for all bit positions until $b-b=0$.

In NN computations, the effects of such perturbations propagate through linear operations and nonlinear activations (e.g. ReLU). 
For a neuron with perturbations in the quantized weights, the error accumulates, shifting the pre-activation value and thus the result of the activation function.

Overall, the robustness of $n$-bit QNNs to perturbations is
determined by the interplay between quantization resolution and classification
margins.
While higher bit widths reduce the magnitude of individual
perturbations, low-bit quantizations have larger relative step sizes and therefore changes in magnitudes in case of perturbations.
It is important to note that reducing the number of bits in uniform quantization does not necessarily improve the NN's tolerance to errors.
We provide empirical evidence of this phenomenon in our evaluation section.

\subsection{Effects of Perturbations in Binarized Neural Networks}
In Binarized Neural Networks (BNNs), the weights and activations are quantized to $\{\pm1\}$.
Due to this, the multiplication, summation, and activation can be computed with
\begin{equation}
    2 \cdot \text{popcount}(\text{XNOR}(\mathbf{W},\mathbf{A})) - \#{bits} > \mathbf{T},
    \label{eq:bnnformula}
\end{equation}
where ${\text{XNOR}}(\mathbf{W},\mathbf{A})$ computes the XNOR of the rows in $\mathbf{W}$ with the columns in $\mathbf{A}$ (analogue to matrix multiplication), popcount counts the number of set bits in the XNOR result, $\#bits$ is the number of bits of the XNOR operands, and $\mathbf{T}$ is a vector of learnable threshold parameters, with one entry for each neuron.
Note that the result of popcount is multiplied by two and then $\#{bits}$ is subtracted, by which the result of popcount is transformed into a value that would have resulted if the computations were performed using the binarization $\{\pm1\}$.
Finally, the comparisons against the thresholds produce again binary values.

To reason about the effects of perturbations, we consider single bit flips.
One bit flip in one weight manifests itself in an increase (0 to 1 flip) or decrease (1 to 0 flip) of the XNOR result by $1$.
As long as the weights or input flips do not cause the pre-activation value to pass the threshold, the activation will not flip, in which case the neuron is bit error tolerant. 

Therefore, the bit error tolerance of an hidden layer neuron $j$ depends on the margin
\begin{equation} \label{eq:Tofy}
  M_{x,j} = \left|s_{x,j} - t_j\right|
\end{equation}
between the pre-activation value ($s_{x,j}$) and the threshold ($t_j$).
With each bit flip $s_{x,j}$ may get closer to $t_j$ and may finally flip the output of the activation if $t_j$ is passed.
A detailed proof for the exact number of bit flips is given in~\cite{buschjaeger:2021}.

An analysis of neuron-based bit error tolerance has been conducted in~\cite{buschjaeger-towards:2020}, showing the relation of the metric in~\ref{eq:Tofy} to the bit error tolerance of BNNs.
Using this metric for optimizing bit error tolerance has been reported to be unsuccessful.
In short, based on the evaluations, the reason given for the negative result is that bit flips of neuron outputs can only affect the BNN prediction if the effect of bit flips reach the output layer and lead to a change of the predicted class.
Focusing on applying the notion of margin to the output layer lead to favorable results~\cite{buschjaeger:2021}, as is also the focus of this work.

In summary, in BNNs, the position of the bit flip does not matter. 
For one bit flip in the weight, the XNOR result always is increased or decreased by $1$.
Furthermore, due to the fact that weights and activations are binarized, the effect of bit flips is bounded.
If the binary output of the neurons do not flip, then the bit flips may not have any impact in the BNN at all. 

\section{Evaluations}
\label{sec:experiments}

The experimental evaluation is structured as follows:
First, the overall experimental setup is described in Section~\ref{subsec:expsetup}. 
Next, the experimental results obtained using the margin-based cross-entropy loss (MCEL) are presented in Section~\ref{subsec:expresults}.
This is followed by an analysis of the margin evolution during training in Section~\ref{subsec:marginevo}.

\subsection{Experiment Setup}
\label{subsec:expsetup}

\begin{table}[t]
  \centering
    \caption{
  {Datasets used for experiments.}}
  \scalebox{1}{
  \begin{tabular}{|l|l|l|l|l|}
  \hline
  Name         & \# Train & \# Test & \# Dim & \# classes       \\ \hline
  FashionMNIST (VGG3) & 60000 & 10000 & (1,28,28) & 10 \\
  SVHN (VGG7)         & 73257    & 26032   & (3,32,32)  & 10  \\   
  CIFAR10 (MobileNetV2)     & 50000    & 10000   & (3,32,32)   & 10  \\
  Imagenette (ResNet18)     & 9470    & 3925   & (3,64,64)   & 10  \\
  \hline
  \end{tabular}
  }
  \label{tab:datasets}
\end{table}

\begin{table*}[t]
  \centering
    \caption{
  {
  QNNs with fully connected (FC), convolutional (C), and maxpool (MP) layers. SCB: Skip-connection block. IRB: Inverted residual block. AAP: Adaptive average pool. Convolutional layers are followed by batch normalization layers, except output layers. EP: epochs, BS: batch size, LR: learning rate, SS: step size, Gamma: decay factor (multiplied with LR every SS epochs). In brackets are the hyperparameters for the BNN models.}}
  \scalebox{1}{
  \begin{tabular}{|l|l|l|l|l|l|l|}
  \hline
  Name                                   & Architecture & EP & BS & LR & SS & Gamma  \\ \hline
  VGG3 & In $\to$ C64 $\to$ MP2  $\to$ C64 $\to$ MP2 $\to$ FC2048  $\to$ FC10 & 100 & 256 & $10^{-3}$ & 10 & 0.5\\[1mm]
  VGG7 & In $\to$ C64  $\to$ C64 $\to$ MP2 $\to$ C128  $\to$ C128 $\to$ MP2 & 50 (100) & 256 & $10^{-3}$ & 5 (10) & 0.25 (0.5)  \\
  & \phantom{In} $\to$ C256  $\to$ C256 $\to$ MP2 $\to$ FC512  $\to$ FC10 & & & & & \\
  MobileNetV2 &
  In $\to$ C32 $\to$ IRB(t1,c32) & 100 & 128 & $10^{-3}$ & 10 & 0.5\\
  & \phantom{In} $\to$ IRB(t6,c64,s2) $\to$ IRB(t6,c96) & & & & &\\
  & \phantom{In} $\to$ C256 $\to$ AAP $\to$ FC10 & & & & & \\

  ResNet18 & In $\to$ C64 $\to$ SCB64 $\to$ SCB128 $\to$ SCB256 & 100 & 128 & $10^{-3}$ & 10 & 0.5\\
  & \phantom{In} $\to$ SCB512  $\to$ AAP $\to$ FC10 & & & & & \\
  \hline
  \end{tabular}
  }
  \label{tab:expsetting}
\end{table*}


The aim in this Section is to explore and evaluate the MCEL across multiple models, datasets, and quantization settings commonly applied in the embedded systems domain, and compare it to the state of the art methods.

To assess the error tolerance of the NNs in a general way, we inject bit flips into the weights, with varying probability, depending on the NN's error tolerance capability, with different bit error rates (BERs). 
This matches the assumptions in recent studies about approximate memories and serves as a model for approximate computing.
We assume that the probability for 0 to flip to 1 is the same as the probability for 1 to flip to 0.
For efficient bit flip injection, we use the framework in 
\url{https://github.com/myay/BNN-QNN-ErrorEvaluation},
which uses highly optimized GPU-kernels for efficient bit-level manipulations. 

We adopt quantization-aware training (QAT) throughout this work, which is also implemented in the framework above.
QAT is performed for all quantization schemes.
As the goal is to isolate and study the mechanisms underlying bit error tolerance in QNNs and find out the effects on the number of encoding bits, we also apply QAT for 8-bit NNs, for which Post-Training-Quantization (PTQ) is typically applied.
QAT ensures that the network is trained and evaluated under identical quantization constraints for all quantization schemes, avoiding any confounding effects.
We therefore consider QAT a more suitable experimental setting for analyzing robustness and margin-based behavior under bit errors.

To compare our method MCEL to a state-of-the-art baseline, we employ the standard cross-entropy loss as implemented in the official PyTorch framework.
For our proposed MCEL method, we augment the loss computation with a tanh-based logit clipping operation  (Equation~\ref{eq:tanhclip}) and apply an explicit margin by subtracting a fixed value $m$ from the logit corresponding to the ground-truth class (Equation~\ref{eq:margin-logits}).
Importantly, the margin modification is applied only within the loss computation and does not alter the NN outputs during inference or training forward passes.
As a result, the NN does not observe the margin directly and cannot satisfy the margin constraint through logit shifting or rescaling.

In our evaluations, we use moderately difficult prediction tasks, with a small VGG3 model and relatively large VGG7, MobileNetV2, and ResNet18 models.
The NNs are up-to-date, suitably sized (not overparametrized) and capable models tailored for resource-constrained inference on the edge.
We use the datasets Fashion, SVHN, CIFAR10, and Imagenette (a subset of ImageNet, scaled to $64 \times 64$ in our work), see Table~\ref{tab:datasets}, inspired by the experiment setups in~\cite{buschjaeger:2021,mohaghaddas:2023}.
Our models (Table~\ref{tab:expsetting}) are quantized architectures of VGG~\cite{simonyan:2014}, MobileNetV2~\cite{sandler:2018}, and ResNet~\cite{he:2016} models, adapted for the above datasets. 
In the case of 2-,4-, and 8-bit QNNs, we use the ReLU activation function.
In the case of BNNs, we use the weakest variant, with binarized weights and binarized activations, which are the hardest to train. 
We use Adam for optimization in all cases.
We use quantization-aware training and quantize the weights, i.e. we quantize the weights during inference and training according to Equation~\ref{eq:quantization}.
Note that we emphasize robustness analysis rather than focusing on fine-tuning for peak accuracy. 
    

Note that preserving the notion of meaningful margins requires the network logits to remain within the approximately linear region of the tanh function, where
$L \cdot \tanh\!\left(\frac{z}{L}\right) \approx z$.
To satisfy this condition, the raw output logits of the neural network are scaled by a constant factor such that their magnitudes are close to unity.
With this scaling, dividing the logits by $L = 100$ ensures that they lie well within the linear regime of the tanh function, thereby preserving relative logit differences and enabling effective margin enforcement.

\subsection{Experiment Results}
\label{subsec:expresults}

\begin{figure*}[!tb]
    \begin{center}
      \begin{subfigure}[b]{0.34\textwidth}
        \includegraphics[width=1\textwidth]{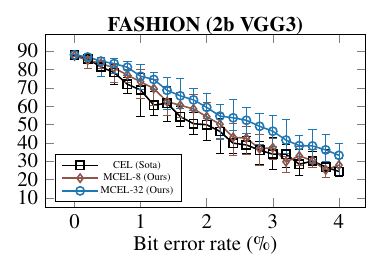}
      \end{subfigure}%
      \begin{subfigure}[b]{0.34\textwidth}
        \includegraphics[width=1\textwidth]{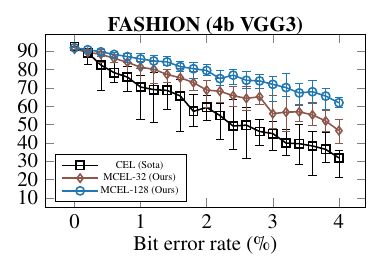}
      \end{subfigure}%
      \begin{subfigure}[b]{0.34\textwidth}
        \includegraphics[width=1\textwidth]{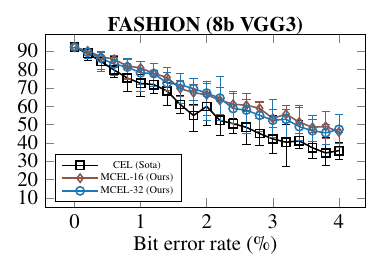}
      \end{subfigure}%
      \vspace{-0.25cm}
      \\
      \begin{subfigure}[b]{0.34\textwidth}
        \includegraphics[width=1\textwidth]{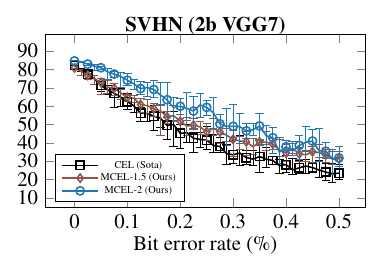}
      \end{subfigure}%
      \begin{subfigure}[b]{0.34\textwidth}
        \includegraphics[width=1\textwidth]{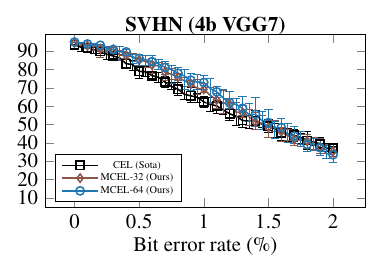}
      \end{subfigure}%
      \begin{subfigure}[b]{0.34\textwidth}
        \includegraphics[width=1\textwidth]{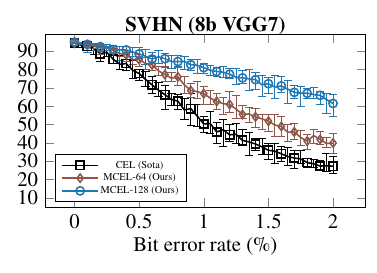}
      \end{subfigure}%
      \vspace{-0.25cm}
      \\
      \begin{subfigure}[b]{0.34\textwidth}
        \includegraphics[width=1\textwidth]{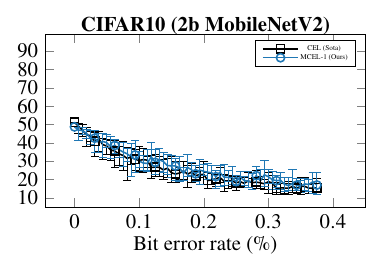}
      \end{subfigure}%
      \begin{subfigure}[b]{0.34\textwidth}
        \includegraphics[width=1\textwidth]{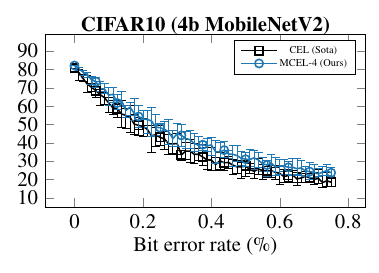}
      \end{subfigure}%
      \begin{subfigure}[b]{0.34\textwidth}
        \includegraphics[width=1\textwidth]{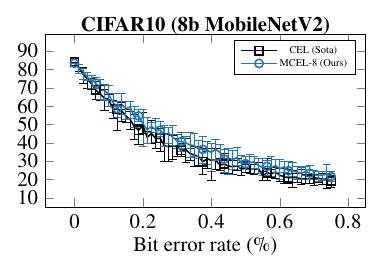}
      \end{subfigure}%
      \vspace{-0.25cm}
      \\
      \begin{subfigure}[b]{0.34\textwidth}
        \includegraphics[width=1\textwidth]{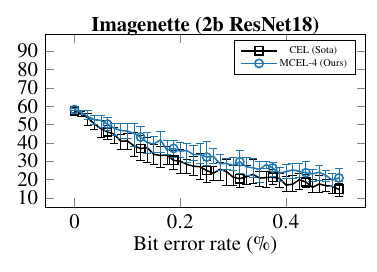}
      \end{subfigure}%
      \begin{subfigure}[b]{0.34\textwidth}
        \includegraphics[width=1\textwidth]{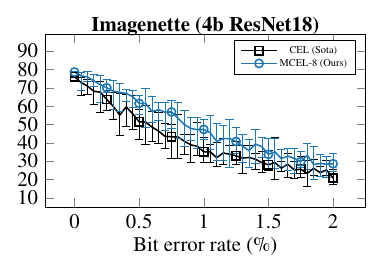}
      \end{subfigure}%
      \begin{subfigure}[b]{0.34\textwidth}
        \includegraphics[width=1\textwidth]{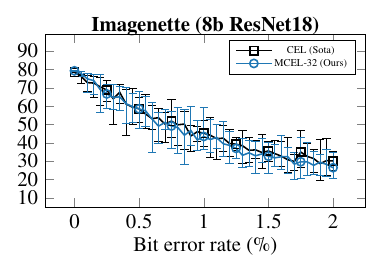}
      \end{subfigure}%
      \\
      \begin{subfigure}[b]{0.34\textwidth}
        \includegraphics[width=1\textwidth]{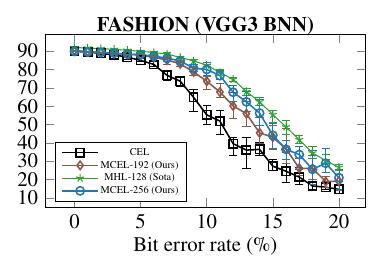}
      \end{subfigure}%
      \begin{subfigure}[b]{0.34\textwidth}
        \includegraphics[width=1\textwidth]{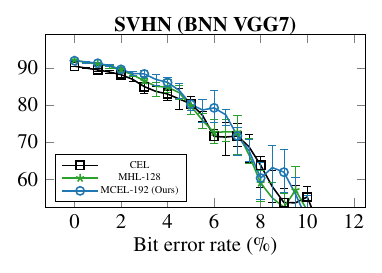}
      \end{subfigure}%
      \caption{Accuracy over bit error rate for 2-, 4-, and 8-bit QNNs, as well as BNNs. The error rate is specified under a given bit flip injection rate on the x-axes. CEL is the state of the art (SOTA), MCEL is our proposed method. We do not perform any bit flip injection during training, we only inject errors during inference to assess error tolerance.}
      \label{fig:celvsmcel-qnns}
    \end{center}
\end{figure*}


\begin{figure}[h!]
    \begin{center}
      \begin{subfigure}[b]{0.5\columnwidth}
        \includegraphics[width=0.965\textwidth]{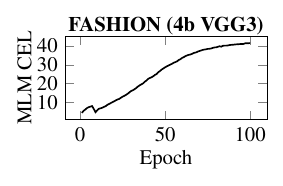}
      \end{subfigure}%
      \begin{subfigure}[b]{0.5\columnwidth}
        \includegraphics[width=1\textwidth]{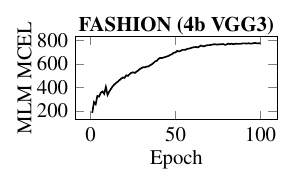}
      \end{subfigure}%
      \vspace{-0.4cm}
      \\
      \begin{subfigure}[b]{0.5\columnwidth}
        \includegraphics[width=0.935\textwidth]{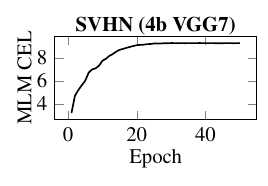}
      \end{subfigure}%
      \begin{subfigure}[b]{0.5\columnwidth}
        \includegraphics[width=0.975\textwidth]{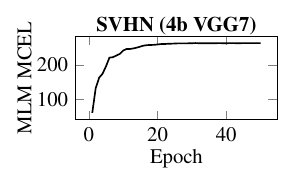}
      \end{subfigure}%
      \vspace{-0.4cm}
      \\
      \begin{subfigure}[b]{0.5\columnwidth}
        \includegraphics[width=1.05\textwidth]{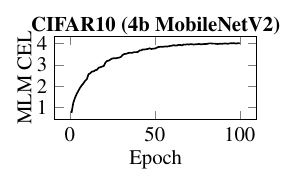}
      \end{subfigure}%
      \begin{subfigure}[b]{0.5\columnwidth}
        \includegraphics[width=1.05\textwidth]{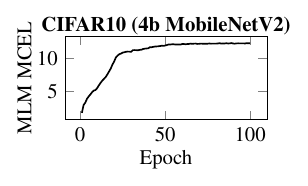}
      \end{subfigure}%
      \vspace{-0.4cm}
      \\
      \begin{subfigure}[b]{0.5\columnwidth}
        \includegraphics[width=1\textwidth]{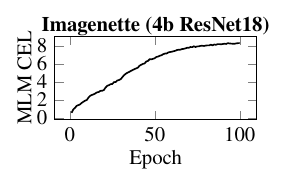}
      \end{subfigure}%
      \begin{subfigure}[b]{0.5\columnwidth}
        \includegraphics[width=1.075\textwidth]{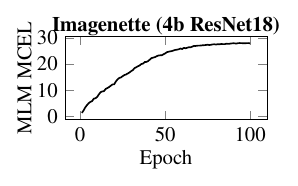}
      \end{subfigure}%
      \\
      \vspace{-0.4cm}
      \begin{subfigure}[b]{0.5\columnwidth}
        \includegraphics[width=1\textwidth]{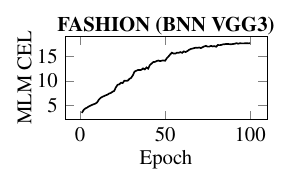}
      \end{subfigure}%
      \begin{subfigure}[b]{0.5\columnwidth}
        \includegraphics[width=1.075\textwidth]{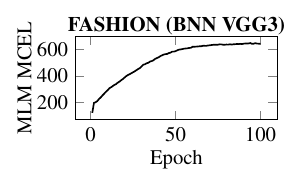}
      \end{subfigure}%
      \\
      \vspace{-0.4cm}
      \begin{subfigure}[b]{0.5\columnwidth}
        \includegraphics[width=0.975\textwidth]{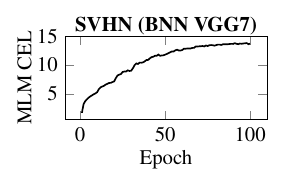}
      \end{subfigure}%
      \begin{subfigure}[b]{0.5\columnwidth}
        \includegraphics[width=1\textwidth]{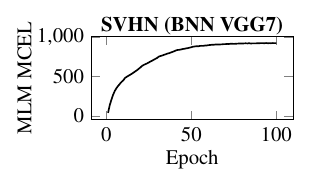}
      \end{subfigure}%
      
      \vspace{-0.25cm}
      \caption{Evolution of the margins between the highest and the second highest logit the NN returns (i.e. MLM for Mean Logit Margin) during training. Left column: CEL. Right column: MCEL. QNNs with 4-bit quantization and BNNs shown as examples. Y-axis: Average of top2 margins (see Equation~\ref{eq:top2} for one margin and Figure~\ref{fig:logit-margin-illustration} for an illustration) over all per-training-sample top2 margins of one epoch.}
      \vspace{-0.5cm}
      \label{fig:marginevo}
    \end{center}
\end{figure}

Figure~\ref{fig:celvsmcel-qnns} summarizes the experimental results for QNNs and BNNs across datasets of varying difficulty, multiple architectures, and different quantization schemes.
Overall, models trained with the proposed MCEL consistently achieve higher robustness to bit errors, measured as classification accuracy under increasing bit error rates, compared to models trained with the standard cross-entropy loss.
In the figure, the blue curves correspond to MCEL-trained models.
For example for FashionMNIST with 4b, the accruacy achieved by MCEL is 15.32\% higher than for CEL for a bit error rate of 1\%. 
Other similar potential operating points can be identified in the plots.  

For all experiments, we evaluate margin values in powers of two up to 128 and $m = 192$, which corresponds to the near maximum relative logit separation achievable with $\frac{m}{2L}$ for $L = 100$, as used throughout this study.
This bounded and interpretable parameterization makes the selection of suitable margin values straightforward.
The highest bit error tolerance is achieved with the margin values reported in the figure legends.
Increasing the margin beyond these values does not yield additional robustness gains and often degrades nominal accuracy.
For margin values exceeding $m > 200$, no measurable difference compared to $192$ or $m = 200$ is observed, except for BNNs with FashionMNIST.


The most challenging experimental configuration for MCEL is Imagenette evaluated with a ResNet18 architecture using 2-bit quantized weights.
In this case, multiple limiting factors coincide.
First, Imagenette images are downscaled to a reduced spatial resolution, significantly reducing the available visual information.
Second, the use of 2-bit weights severely constrains the representational capacity of the network.
Finally, MCEL enforces an explicit margin in the output logits, further restricting the feasible solution space.
As a result, the network is constrained from multiple directions simultaneously, limiting its ability to achieve both high classification accuracy and large confidence margins.
Empirically, this manifests as slow or unstable convergence, with accuracy saturating at approximately $60\%$, even when trained using the standard cross-entropy loss.

The impact of MCEL further depends on the employed quantization scheme.
For 2- and 4-bit quantized neural networks, MCEL consistently provides substantial robustness improvements.
In contrast, the performance gap between MCEL and standard cross-entropy loss is smaller for 8-bit quantization.
For Imagenette with 8-bit weights, MCEL struggles to achieve high levels of bit error tolerance, indicating that the effectiveness of margin-based robustness mechanisms diminishes as quantization noise becomes less dominant.

As shown in Figure~\ref{fig:celvsmcel-qnns} (last row), MCEL is also effective for BNNs.
For FashionMNIST with VGG3, MHL (modified hinge loss from ~\cite{buschjaeger:2021}) slightly outperforms MCEL, although the performance difference is small.
In this case, MCEL with $m=256$ leads to higher error tolerance than $m=192$, indicating high potential for error tolerance with this NN and dataset combination.
In contrast, for SVHN with VGG7, MCEL outperforms MHL, demonstrating that margin-based optimization at the logit level generalizes beyond quantized multi-bit NNs.
For the BNNs models, we adopt the VGG3 and VGG7 architectures from~\cite{buschjaeger:2021} to enable a direct comparison between the modified hinge loss (MHL) and MCEL.
Constructing competitive BNN variants of more complex architectures such as ResNet or MobileNetV2 remains an open research problem and typically requires additional techniques, including application-specific architectural modifications, neural architecture search, or ensemble methods~\cite{shankar:2024,phan:2020}.
Restricting the comparison to established BNN models ensures comparability.

Taken together, the results on QNNs and BNNs empirically confirm the theoretical predictions underlying MCEL and demonstrate that the method behaves as intended across a wide range of datasets, architectures, and quantization levels.

\subsection{Margin Evolution}
\label{subsec:marginevo}

The evolution of the margin between the highest and second-highest logits produced by the NNs, referred to as the \emph{Mean Logit Margin (MLM)}, is shown in Figure~\ref{fig:marginevo}.
As representative examples, we consider the NN models trained with 4-bit weight quantization.
For each training epoch, the MLM is computed by averaging the top-2 logit margins over all training samples processed during that epoch.
The top-2 margin for an individual sample is defined in Equation~\ref{eq:top2} and an intuitive illustration is in Figure~\ref{fig:logit-margin-illustration}.

For 4-bit QNNs with FashionMNIST and SVHN, the MCEL exhibits an MLM that is approximately twenty times larger than that obtained with the standard cross-entropy loss.
For CIFAR10 and Imagenette, the MCEL exhibits an approximately 3 times higher MLM than CEL.
We also observed in Figure~\ref{fig:celvsmcel-qnns} that in these two cases, the achieved error tolerance is lower compared to the cases of Fashion or SVHN.
For the BNNs, the MLM difference between MCEL and CEL is the largest, the factor is approximately 30 for Fashion and 60 for SVHN.

These observations confirm that MCEL effectively enforces larger output-layer margins, as intended.
In contrast, classical CEL primarily ensures that the logit corresponding to the correct class exceeds competing logits, but does not explicitly encourage a substantial separation between them.



\section{Discussion}

A natural point of comparison for our approach is the modified hinge loss (MHL) proposed in~\cite{buschjaeger:2021}, which was shown to increase bit error tolerance in BNNs without requiring error injection during training.
We investigated whether a similar effect could be achieved for quantized neural networks (QNNs) by applying the MHL in our experimental setup.
However, despite extensive hyperparameter tuning, we consistently observed that training QNNs with the MHL resulted in significantly lower classification accuracy compared to standard cross-entropy loss, while error tolerance did not improve in a meaningful way. 
The underlying reasons for this discrepancy between BNNs and QNNs remain unclear and warrant further investigation. 
We hypothesize that structural differences between BNNs and QNNs, such as the use of ReLU activations and multi-level weight representations in QNNs, fundamentally alter the optimization dynamics and error propagation behavior, limiting the effectiveness of hinge-based objectives.
In contrast, our proposed MCEL formulation provides an interpretable margin parameter through the relative logit separation $RLS = \frac{m}{2L}$, enabling principled tuning of robustness targets without costly hyperparameter search. 
Moreover, unlike the hard clipping behavior inherent to MHL, the smooth tanh-based clamping used in MCEL preserves gradient continuity and leads to more stable optimization, particularly in low-precision regimes where training dynamics are already fragile.

While the proposed margin-based framework proves effective for classification tasks with explicit logits, it is subject to certain limitations. 
Our analysis assumes a discrete class setting in which prediction confidence can be expressed through a linear ordering of output logits and where classification margins are well defined.
As a result, the proposed robustness mechanism does not directly extend to non-logit-based models, such as generative architectures, in which class-wise logits are absent or lack a clear semantic interpretation. 
Similarly, models with structured outputs, such as sequences, or graphs, present additional challenges, as outputs are no longer discrete classes and the notion of a margin depends on the geometry of the output space and the encoding of target structures.
Extending margin-based robustness concepts to these settings, and identifying suitable analogues of logit separation for non-classification tasks, is an important direction for future research.



\section{Conclusion}

We establish a direct link between robustness and NN output-layer margins and exploit this insight to derive our proposed margin-cross-entropy loss (MCEL).
Extensive experiments across multiple datasets, architectures, and quantization schemes demonstrate that MCEL substantially improves robustness to bit errors while maintaining prediction accuracy.
Moreover, the method is simple to implement, computationally efficient, and can be integrated into existing training pipelines as a drop-in replacement for standard cross-entropy loss.

Beyond its empirical effectiveness, MCEL highlights a broader shift in how error tolerance in NNs can be addressed.
Rather than relying on costly and increasingly unscalable training-time error injection, our results show that focusing on the underlying mechanisms of robustness leads to interpretable, explainable, and tunable optimization methods.
We view this work as a first step toward generally robust NNs trained with minimal modifications to the training procedure, paving the way for scalable robustness on future approximate computing platforms.

\bibliography{references}
\bibliographystyle{ieeetr}
\end{document}